\documentclass[letterpaper]{article} 
\usepackage{aaai2026}  
\usepackage{times}  
\usepackage{helvet}  
\usepackage{courier}  
\usepackage[hyphens]{url}  
\usepackage{graphicx} 
\usepackage{natbib}  
\usepackage{caption} 
\usepackage{algorithm}
\usepackage{algorithmic}
\usepackage{amsfonts}
\usepackage{booktabs}       
\usepackage{nicefrac}       
\usepackage{microtype}      
\usepackage{xcolor}         
\usepackage{amsmath}
\usepackage{amssymb}
\usepackage{mathtools}
\usepackage{amsthm}
\usepackage{subfigure}
\usepackage{makecell}
\usepackage[textsize=tiny]{todonotes}
\usepackage{tikz}
\newcommand*\circled[1]{\tikz[baseline=(char.base)]{
            \node[shape=circle,draw,inner sep=1.5pt] (char) {#1};}}
\usepackage{newfloat}
\usepackage{listings}
\DeclareCaptionStyle{ruled}{labelfont=normalfont,labelsep=colon,strut=off} 
\floatstyle{ruled}
\newfloat{listing}{tb}{lst}{}
\floatname{listing}{Listing}
\title{Large Connectome Model: An fMRI Foundation Model of Brain Connectomes Empowered by Brain-Environment Interaction in Multitask Learning Landscape}
\author{
    Ziquan Wei\textsuperscript{\rm 1},
    Tingting Dan\textsuperscript{\rm 2},
    Guorong Wu\textsuperscript{\rm 2,1}\thanks{Corresponding author.}
}
\affiliations{
    \textsuperscript{\rm 1}Department of Computer Science, \textsuperscript{\rm 2}Department of Psychiatry\\
  University of North Carolina at Chapel Hill\\
  Chapel Hill, NC 27599 USA \\
    \texttt{ziquanw@\{cs,email\}.unc.edu;\{Tingting\_Dan,grwu\}@med.unc.edu}
}

\begin{document}

\maketitle

\begin{abstract}
A reliable foundation model of functional neuroimages is critical to promote clinical applications where the performance of current AI models is significantly impeded by a limited sample size. 
To that end, tremendous efforts have been made to pretraining large models on extensive unlabeled fMRI data using scalable self-supervised learning. 
Since self-supervision is not necessarily aligned with the brain-to-outcome relationship, most foundation models are suboptimal to the downstream task, such as predicting disease outcomes.
By capitalizing on rich environmental variables and demographic data along with an unprecedented amount of functional neuroimages, we form the brain modeling as a multitask learning and present a scalable model architecture for (i) multitask pretraining by tokenizing multiple brain-environment interactions (BEI) and (ii) semi-supervised finetuning by assigning pseudo-labels of pretrained BEI.
We have evaluated our foundation model on a variety of applications, including sex prediction, human behavior recognition, and disease early diagnosis of Autism, Parkinson's disease, Alzheimer's disease, and {Schizophrenia}, where promising results indicate the great potential to facilitate current neuroimaging applications in clinical routines.
\end{abstract}


\section{Introduction}
\label{sec:intro}

A scalable foundation model dedicated to brain activity is critical to discovering the enigma of human cognition and promoting clinical applications from large-scale neuroimaging data. The topic of the brain foundation model is under exploration since BrainLM \cite{ortega2023brainlm} via masked autoencoder. Previous works formulate this problem by mimicking natural language or image foundation models as self-regression or masking strategies, that is learning the raw signal reconstruction, e.g., Masked Autoencoder (MAE) in \cite{ortega2023brainlm}, Joint-Embedding Predictive Architecture (JEPA) in \cite{dong2024brain}, and other masking methods \cite{wen2023graph,yang2024brainmass}. However, as revealed by Meta-matching \cite{he2022meta}, the brain connectomes share similar features across arbitrary phenotypic traits (hereafter shortened to `phenotypes') such as age and lifestyle. The multitask learning for phenotypic prediction can hence act as the foundation objective related to neuroscience interests. Furthermore, although existing brain foundation models \cite{ortega2023brainlm,dong2024brain} take the raw signal as input, the vast majority of related works \cite{cui2022braingb,said2023neurograph,ding2024machine,wei2024neuropath} suggest that brain connectomes as the input makes more accurate predictions for clinical applications.

\begin{figure}
    \centering
    \includegraphics[width=\linewidth]{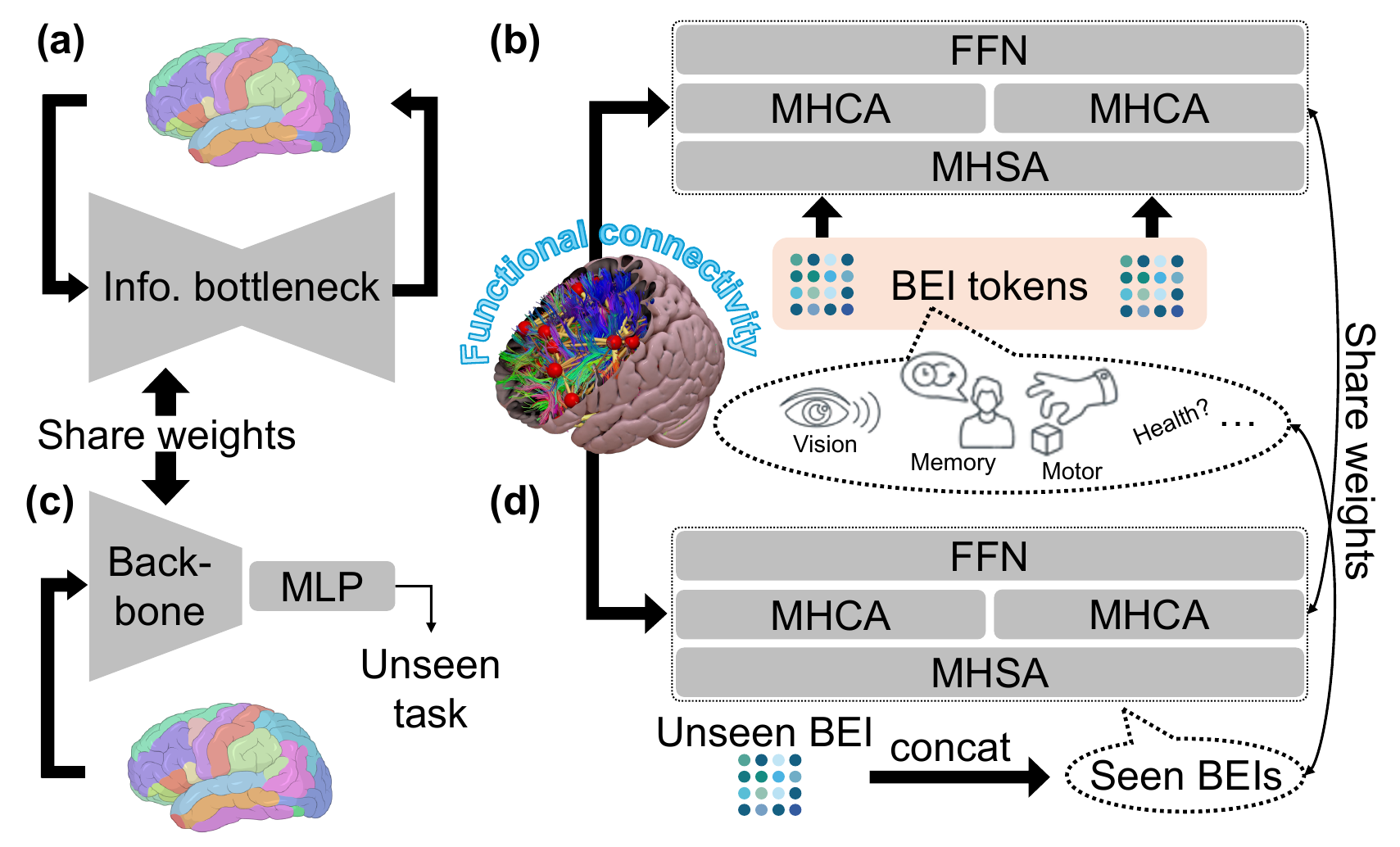}
    \vspace{-2em}
    \caption{
    Learning strategies of previous brain foundation models and LCM. The pretraining in \textbf{(a)} previous brain foundation models is a reconstructive representation learning based on the information (info.) bottleneck, while \textbf{(b)} in LCM it is a multitask learning for multiple brain-environment interactions (BEI) token embeddings by a Transformer decoder, where MHSA is multi-head self-attention, MHCA is multi-head cross-attention, and FFN is a feedforward network. The finetuning in \textbf{(c)} previous studies is training a relatively small head, e.g., a multilayer perceptron (MLP), for the downstream task. \textbf{(d)} LCM finetunes the BEI tokens along with new tokens representing the downstream task.
    }
    \vspace{-1em}
    \label{fig1:intro}
\end{figure}

The self-regressive methodology, although it demonstrated excellent applications such as GPTs for natural language \cite{achiam2023gpt}, should not be the only choice for brain foundation models. As shown in Fig. \ref{fig1:intro}a, the purpose of previous brain foundation models is the same as the image/language to reconstruct the raw signal from its masked version via a bottleneck or transformer encoder architecture. Unlike image/language that is not naturally labeled, brain fMRI has demographics and phenotypes, e.g., age, biological sex, cognitive state, etc, which are commonly recorded during data acquisition. However, the challenging heterogeneity in multi-phenotype learning risks the robustness of current brain foundation models due to the predictive head is relatively lightweight (Fig. \ref{fig1:intro}c). According to the nature of fMRI, which always has non-imaging records, a scalable architecture of the brain foundation model is necessary to involve multitask learning from rich environmental information relevant to brain cognition.

Demographics and phenotypes that generate a diverse range of brain-environment interactions (BEI) have been found inter-correlated to each other, given the brain connectomes \cite{he2022meta}. Even without finetuning, the classical regression model can outperform the trained version after a basic meta-matching between phenotypes. This motivates us to develop and release a brain foundation model powered by BEI multitask learning, as illustrated in Fig. \ref{fig1:intro}b. Given that cognition relevant BEIs represent the brain function used for diagnosing and other downstream tasks, the downstream outcome can be decoded from the BEI token embeddings as shown in Fig. \ref{fig1:intro}d, where the cross-attention is computed between brain connectome feature and the tokenized BEI. This utilizes the findings in \cite{he2022meta} that seen and unseen cognition relevant phenotypes can be meta-matched. The self-attention in Fig. \ref{fig1:intro} communicates information learned from pretrained BEIs in downstream tasks, enhancing the generality and robustness of the model.

To this end, this work presents three main contributions. (1) A new brain foundation model architecture is proposed to cooperate with multitask pretraining and semi-supervised finetuning on rich BEIs. (2) The \textit{{largest}} connectome model (LCM) for brain fMRI is designed and released along with the pretrained weights based on large scale data ($n=$10,036). (3) Experiments on 8 fMRI datasets evaluate the performance of LCM on sex prediction, human behavior recognition, and disease early diagnosis of Autism, Parkinson’s disease, Alzheimer’s disease, and {Schizophrenia}, in terms of scalability, pretraining, and finetuning.


\begin{figure}[t]
    \centering
    \includegraphics[width=\linewidth]{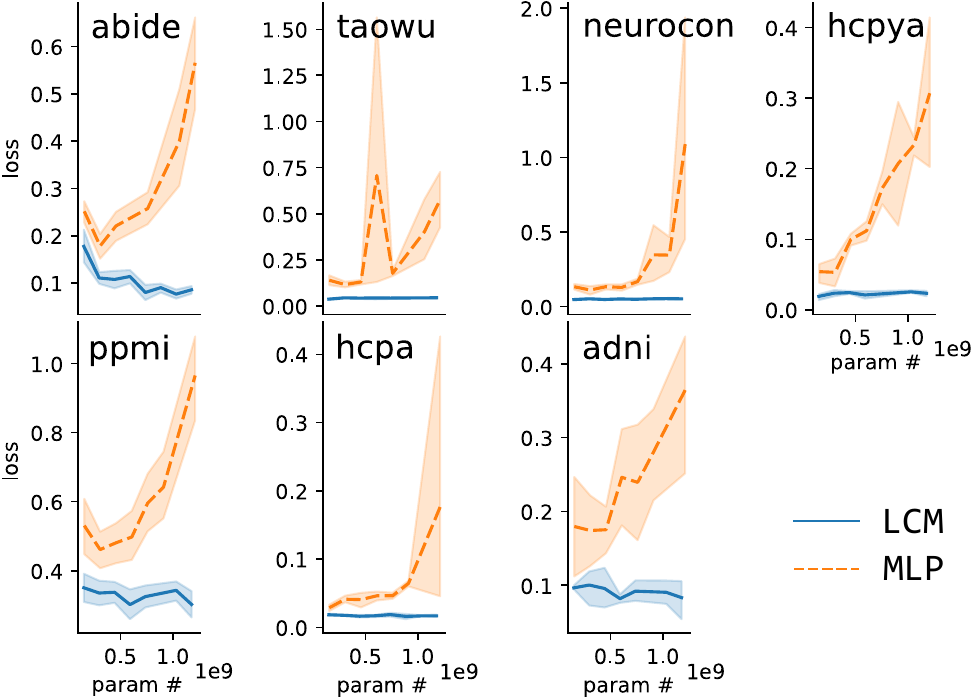}
    \vspace{-1.5em}
    \caption{Scalability is demonstrated by model size vs. training loss, where the training is supervised by arbitrary non-brain-imaging phenotypes as BEIs in our multitask learning.}
    \vspace{-1em}
    \label{fig:loss-nlayer}
\end{figure}

\section{Preliminaries}
\label{sec:pre}

The dynamic signal of brain functional MRI is a blood-oxygen dependent level (BOLD). BOLD signal, which is influenced by a mixture of factors and distorted by non-neuronal fluctuations, has a relatively low signal-to-noise ratio (SNR) \cite{caballero2017methods}. Brain connectomes, on the other hand, increase the SNR in raw signals by representing brain activity via the Pearson correlation coefficient, which is also called functional connectivity (FC) and is defined as follows: 
\begin{equation}\label{eq:fc}
    FC_{ij} = \frac{\sum(\mathbf{x}_{it}-\bar{\mathbf{x}_{i}})(\mathbf{x}_{jt}-\bar{\mathbf{x}_{j}})}{\sqrt{\sum(\mathbf{x}_{it}-\bar{\mathbf{x}_{i}})^2\sum(\mathbf{x}_{jt}-\bar{\mathbf{x}_{j}})^2}},
\end{equation}
where $\mathbf{x}\in \mathbb R^{N\times T}$ refers to the BOLD signal with $N$ and $T$ denoting the number of brain regions and time points, respectively.

Various works have shown superior performance using FC compared to the raw BOLD signal for downstream applications. For example, benchmark papers \cite{cui2022braingb,said2023neurograph,ding2024machine} evaluated the performance by using the BOLD or the correlation as the input, and the BOLD signal has consistently demonstrated lower accuracy. In addition, both static FC and dynamic FC (using the sliding window technique) outperform the BOLD signal \cite{wei2024neuropath}.

\begin{figure}[t]
    \centering
    \includegraphics[width=\linewidth]{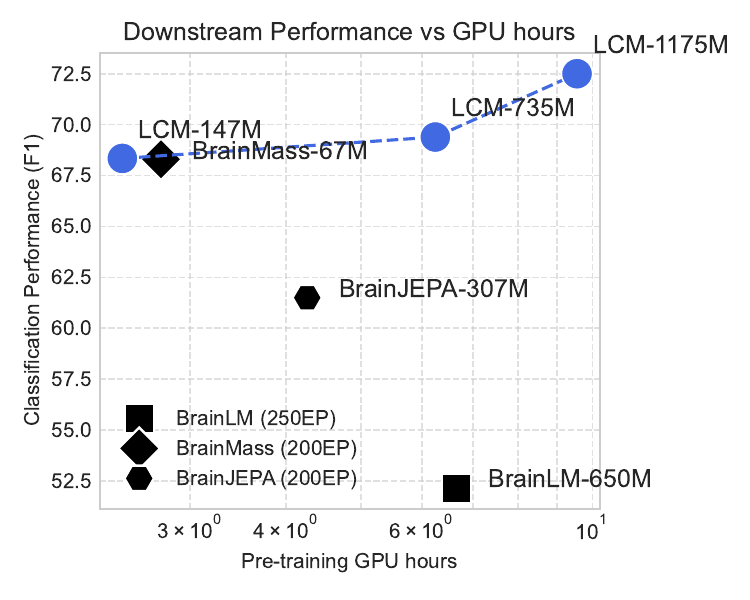}
    \vspace{-2em}
    \caption{Our LCM surpasses other foundation models, demonstrating outstanding efficiency, on our biggest downstream application, ABIDE ($n$=1,025), as an example. Even the smallest LCM (147M), achieves comparable performance while being significantly efficient in both parameters and resource usage.}
    \vspace{-1em}
    \label{fig:f1-gpuhr}
\end{figure}

Therefore, we formulate the problem of the brain foundation model as a pretrained large model dedicated to the brain connectome and coined as the large connectome model (LCM). Inspired by \cite{he2022meta}, LCM is supervised by phenotypic labels, aka. BEI, via multitask learning, which mostly are categorical labels. However, even tremendous efforts have been made for brain connectome and graph classification or regression \cite{ying2021transformers,kan2022brain,chen2022nagphormer,bedel2023bolt,wei2024neuropath}, they, as the encoder, have a limited scalability \cite{wei2024neuropath}. Plenty of theoretical \cite{keriven2022not} and experimental analysis \cite{rusch2023survey} suggest that the reason for that includes over-smoothing and over-squashing. 

\textbf{Scalability analysis:} For brain foundation models, MLP is commonly used as the predictive head \cite{ortega2023brainlm,yang2024brainmass,dong2024brain}. However, as shown in Fig. \ref{fig:loss-nlayer} orange curves, MLP has an exploded training loss when scaling up the parameter amount, where the MLP model is constructed with ReLU activations and residual connections in-between blocks. In this case, we need to think out of the box about the predictive head that has unsatisfactory scalability for the multitask pretraining.

Recently, \cite{paul2024a} proposed a simple framework by replacing the MLP with the Transformer decoder as an interpretable predictive head showing similar or better performance for computer vision tasks. Based on this decoder classifier, our LCM is designed as a decoder-only architecture. In comparison, as shown in Fig. \ref{fig:loss-nlayer} blue curves, LCM shows a better scalability on seven different datasets, where training loss can be lower with more layers used in LCM. This enables LCM to learn from large-scale brain connectome data powered by BEI multitask learning. Consequently, the model efficiency surpasses previous foundation models, which are derived from vision-based encoders, as shown in Fig. \ref{fig:f1-gpuhr}.

\textbf{Related works:} BrainLM \cite{ortega2023brainlm}, to our best knowledge, is the first brain foundation model by applying MAE on BOLD signals. It fills every bit of the fMRI time series can hinder the model’s ability to distinguish between noise and actual signals. However, research \cite{assran2023self} has shown that masked pretraining in generative architectures like MAE often results in suboptimal performance in off-the-shelf evaluations (e.g., linear probing). 
BrainJEPA \cite{dong2024brain}, similarly, framed a new architecture with a different masking strategy JEPA. It handles the suboptimal issues of BrainLM by following the idea of the I-JEPA \cite{assran2023self}. Although their results have shown better performance than linear probing, none of the explicit designs have been added for learning from inter-correlated phenotypes. 
{BrainMass \cite{yang2024brainmass} used a matching objective between pseudo FC matrices by masking BOLD signals. Whilst, it overlooked the phenotypes and demographics that is always assigned with the functional neuroimaging data. }
To this end, we present a scalable large connectome model explicitly supervised by rich environmental variables and demographic data, along with an unprecedented amount of functional neuroimages.







\begin{figure}[t]
    \centering
    \includegraphics[width=\linewidth]{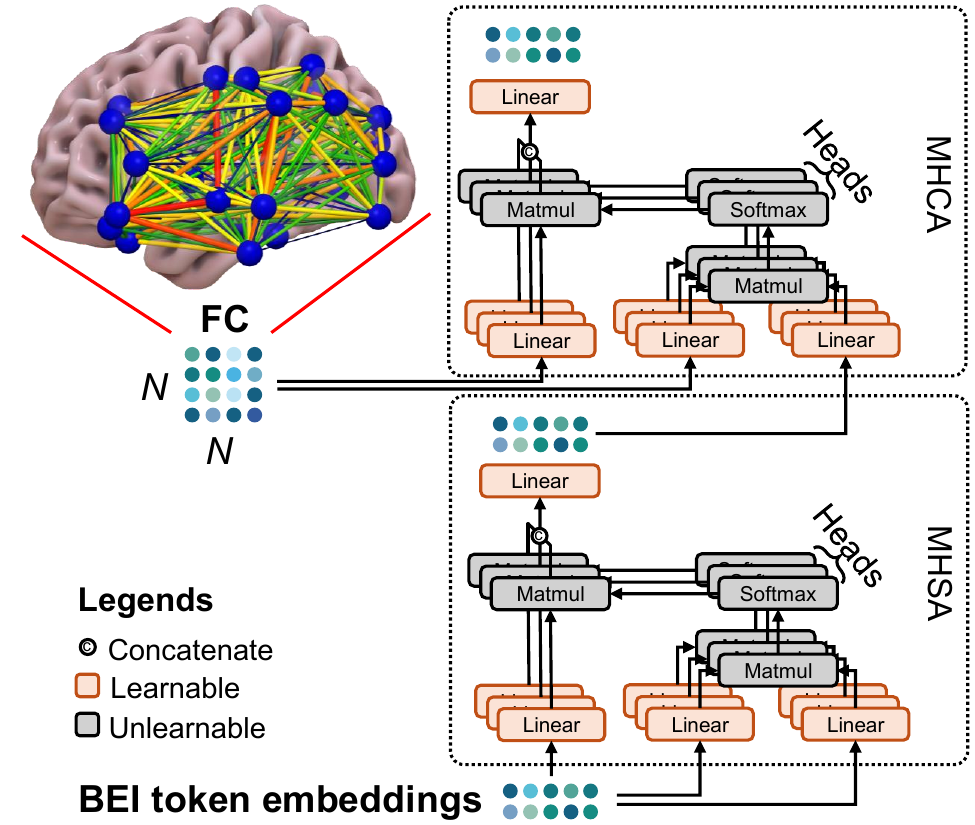}
    \vspace{-1.5em}
    \caption{The architecture of one layer of the LCM.}
    \label{fig:method-arch}
\end{figure}

\section{Methods}\label{sec:method}

Given the FC matrix of fMRI data as defined by Eq. \ref{eq:fc}, LCM takes $FC$ as input and learns from the supervision of multiple non-imaging records of fMRI.

\subsection{{Architecture}}

Inspired by the Transformer-based large models \cite{achiam2023gpt} and \cite{paul2024a}, a decoder-only architecture as shown in Fig. \ref{fig:method-arch} is employed in our LCM.
The token vectors in Fig. \ref{fig:method-arch} refer to $N_{BEI}$ available BEIs in pretraining datasets, which are initialized randomly as the token embedding, denoted by $\mathbf{V}\in\mathbb R^{P\times E}$, where $P=\sum_{i}^{N_{BEI}}\textbf{N}^{class}_i$ and $\textbf{N}^{class}_i$ is the class number of the $i^{th}$ categorical BEI or 1 if the BEI is a continous value. Next, $\mathbf{V}$ is updated by self-attention as follows:
\begin{equation}
    \mathbf{V} =\texttt{Softmax}\left({(\mathbf{V\boldsymbol{\bar\alpha}}_h)(\mathbf{V\boldsymbol{\bar\beta}}_h)^T}/{\sqrt{D}}\right)\left(\mathbf{V\boldsymbol{\bar\gamma}}_h\right),
\end{equation}
where $\boldsymbol{\bar\alpha}_h, \boldsymbol{\bar\beta}_h, \boldsymbol{\bar\gamma}_h\in\mathbb R^{E\times D}$ are learnable parameters of self-attention linear layers shown in Fig. \ref{fig:method-arch}, $h$ is the head index, and $D$ is the hidden channel. 

Suppose $\mathbf{M} \in \mathbb{R}^{N\times N}$ is FC matrix. Cross-attention between $\mathbf V$ and $\mathbf{M}$ is then defined as follows
\begin{equation}
    \mathbf{V} =\texttt{Softmax}\left({(\mathbf{M}\mathbf{\boldsymbol{\hat\alpha}}_h)(\mathbf{V\boldsymbol{\hat\beta}}_h)^T}/{\sqrt{D}}\right)^T\left(\mathbf{M}\mathbf{\boldsymbol{\hat\gamma}}_h\right),
\end{equation}
where $\boldsymbol{\hat\alpha}_h, \boldsymbol{\hat\gamma}_h\in\mathbb R^{N\times D}, \boldsymbol{\hat\beta}_h\in\mathbb R^{E\times D}$ are learnable parameters of cross-attention linear layers shown in Fig. \ref{fig:method-arch}, and $D$ is the hidden channel. Note that the bias in linear layers is omitted in this section for clarity.

This design allows LCM to be easily stacked since each layer updates $\mathbf{V}$ without changing the tensor shape. 

\subsection{Multitask Pretrain and {Semi-supervised Finetune}}

Take categorical BEI as an example, the multitask pretraining is accomplished by
\begin{equation}
L_{cls} = \sum_{i=0}^{N_{BEI}}CELoss(\textbf{V}_{S_i:S_{i+1}},GT_{i}),
\end{equation}
where $S_0=0,S_i=\sum_{j=0}^{i-1}\textbf{N}^{class}_j$ if $i>0$ and $CELoss$ denotes cross-entropy loss. The Mean Squared Error (MSE Loss) is used for regressive BEI tokens. 

Finetuning objectives are the same as pretraining.
Given unseen datasets that have $\hat{N}_{BEI}$ tasks as new BEIs with $\mathbf{\hat{N}}^{class}$ as the vector of class number for each BEI, the finetuning can be easily achieved by concatenating new tokens by updating $\mathbf V\xleftarrow{}[\mathbf V, \mathbf{\hat{V}}]$, $\textbf{N}^{class}\xleftarrow{}[\textbf{N}^{class}, \mathbf{\hat{N}}^{class}]$. For pretrained tokens during finetuning, the pseudo-label is assigned to them with corresponding neuroimaging configurations, e.g., `resting-state' for clinical applications.

\begin{figure*}[t]
    \centering
    \includegraphics[width=.8\linewidth]{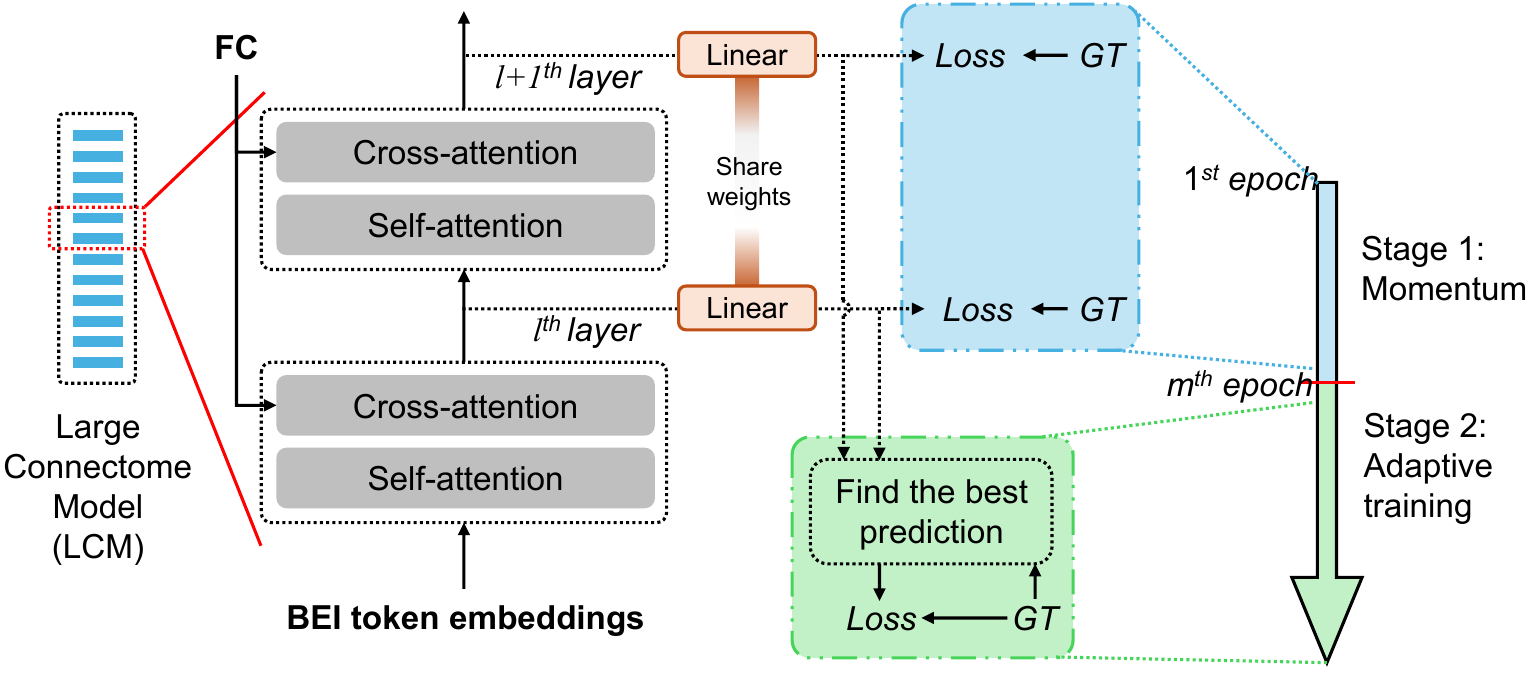}
    \vspace{-1em}
    \caption{Pre-training and finetuning of LCM use a two-stage learning strategy: (1) Getting momentum by computing loss for all layers, and (2) adaptive training for the best layer. Note that ground truth (GT) could be a pseudo-label of the BEI, e.g., subjects are healthy by default in HCP datasets.}
    \vspace{-1em}
    \label{fig:method-train}
\end{figure*}

In case that BEIs differ on the complexity of feature representation, e.g., Parkinson's disease at different stages of treatment might show different symptoms, we propose that LCM predicts each BEI at different layers of the model.
As shown in Fig. \ref{fig:method-train}, the BEI token embeddings from every LCM layer are stored at first. After the computation of LCM, token embeddings from all layers are input into a linear layer: $\mathbb R^{P\times D}\xrightarrow{}\mathbb R^{P\times 1}$ to get various predictions from different layers of LCM. The best prediction, i.e., the layer that has the best predictive score, can be found given the ground truth (GT) during training. The best prediction is then used as the output to get the loss with GT. During testing, the output of the layer that has the highest score is the final output of LCM.

The initialization of our learning strategy is important to have a proper starting point and a correct direction. To account for this, as shown in Fig. \ref{fig:method-train} left part, training the LCM has two stages for both pretraining and finetuning, (1) utilize the average prediction from all layers to update the LCM parameters in the first $m$ epochs, and (2) supervise only the best prediction in the rest of epochs. Namely, stage 1 produces a `momentum' that can push the training of LCM to the correct direction, and hence LCM can achieve a diverse and correct feature representation for different phenotypes in the following stage.



\section{Experiments}
\label{sec:exp}

We evaluate the proposed LCM on 8 datasets including HCP Aging (HCPA), HCP Young Adult (HCPYA), ADNI, PPMI, ABIDE, Taowu, Neurocon, and SZ. Two HCPs contain more than 10,000 scans of brain fMRI from about 1,800 subjects under various cognitive states, depending on resting or tasking. Six disease-related datasets contain about 1,500 subjects under the same resting state but various health status due to different brain-environment interactions.

To comprehensively evaluate and showcase the performance of the proposed LCM, we conduct experiments on both randomly initialized and pretrained models across clinical applications, as well as tasks involving sex, and cognitive state recognition. The fewshot finetuning experiments are also conducted. Pretrained models are finetuned with different ratio of data in the same validation fold to demonstrate performance for real-world clinical applications.

\begin{table*}[t]
    \centering
    \small
    \setlength{\tabcolsep}{2pt}
    \vspace{-1.5em}
    \caption{Finetune LCM with weights learned from various combination of BEIs. Diverse BEIs contribute differently to LCM pretraining, where checkmarks indicate which BEI (\circled{1}: cognitive state, \circled{2}: Alzheimer's, \circled{3}: Parkinson's, and \circled{4}: Autism) are involved, $\dagger$ denotes LCM is not pretrained, $\ddagger$ denotes BrainLM is finetuned on the released model weights, \textbf{Bold} indicates the first ranking place, and \underline{underline} indicates the second.}
    \vspace{-.5em}
    \begin{tabular}{lllll|llllll}
    \toprule
         & \multicolumn{4}{c}{Pretrained on} & \multicolumn{2}{c}{Alzheimer's} & \multicolumn{2}{c}{Parkinson's} & \multicolumn{2}{c}{Autism}\\\cmidrule(lr){2-5}\cmidrule(lr){6-7}\cmidrule(lr){8-9}\cmidrule(lr){10-11}
         & \circled{1} & \circled{2} & \circled{3} & \circled{4} & Accuracy & F1 score & Accuracy & F1 score & Accuracy & F1 score \\
    \midrule
        BrainGNN & & &  &  & 83.48$_{\pm6.99}$ & 78.69$_{\pm8.25}$ & 76.92$_{\pm15.60}$ & 75.24$_{\pm16.43}$ & 62.83$_{\pm2.77}$ & 61.73$_{\pm3.83}$\\
        BolT & & &  &  & 82.89$_{\pm8.66}$ & 77.66$_{\pm10.03}$ & 73.30$_{\pm20.23}$ & 69.93$_{\pm23.98}$ & 68.09$_{\pm3.78}$ & 68.01$_{\pm3.72}$\\
        BNT & & &  &  & 83.56$_{\pm7.47}$ & 78.31$_{\pm12.17}$ & 76.79$_{\pm14.59}$ & 71.99$_{\pm17.45}$ & 70.23$_{\pm3.69}$ & 69.98$_{\pm3.73}$\\
        Graphormer & & &  &   & 82.81$_{\pm7.90}$ & 77.51$_{\pm11.46}$ & 65.89$_{\pm14.50}$ & 61.84$_{\pm17.07}$ & 57.66$_{\pm2.79}$ & 57.00$_{\pm2.92}$\\
        NAGphormer & & &  &   & 80.59$_{\pm6.82}$ & 76.07$_{\pm9.72}$ & 72.16$_{\pm18.01}$ & 67.84$_{\pm19.93}$ & 64.78$_{\pm2.47}$ & 64.60$_{\pm2.73}$\\
        NeuroPath & & &  &   & 82.07$_{\pm6.86}$ & 74.12$_{\pm9.56}$ & 69.79$_{\pm17.43}$ & 65.81$_{\pm19.77}$ & 65.71$_{\pm5.93}$ & 64.64$_{\pm7.36}$\\
        LCM$\dagger$ & & &  &  & {85.04$_{\pm9.30}$} & 79.58$_{\pm13.77}$ & 70.25$_{\pm14.36}$ & 66.10$_{\pm16.62}$ & 69.65$_{\pm5.10}$ & 69.58$_{\pm5.23}$\\
        \midrule
        BrainLM$\ddagger$ & \checkmark & & & & 83.30$_{\pm4.71}$ & 75.79$_{\pm6.63}$ & 50.43$_{\pm19.59}$ & 45.83$_{\pm23.41}$ & 53.25$_{\pm4.00}$ & 51.51$_{\pm5.67}$\\
        BrainLM & \checkmark & \checkmark & \checkmark & \checkmark & 82.56$_{\pm4.01}$ & 75.41$_{\pm6.21}$ & 54.40$_{\pm12.37}$ & 50.37$_{\pm13.25}$ & 49.90$_{\pm4.86}$ & 33.89$_{\pm5.18}$\\
        {BrainMass-SVM} & \checkmark & \checkmark & \checkmark & \checkmark & 82.96$_{\pm5.02}$ & 75.32$_{\pm7.06}$ & 59.21$_{\pm21.68}$ & 51.25$_{\pm25.07}$ & 64.49$_{\pm1.59}$ & 64.56$_{\pm1.62}$\\
        {BrainMass-MLP} & \checkmark & \checkmark & \checkmark & \checkmark & 82.96$_{\pm5.02}$ & 75.32$_{\pm7.06}$ & 73.78$_{\pm17.95}$ & 70.20$_{\pm21.02}$ & 68.48$_{\pm4.50}$ & 68.30$_{\pm4.70}$\\
        {Brain-JEPA} & \checkmark & \checkmark & \checkmark & \checkmark & \underline{86.02$_{\pm3.49}$} & 80.20$_{\pm4.95}$ & \underline{82.09$_{\pm9.58}$} & 77.16$_{\pm12.34}$ & 63.84$_{\pm0.82}$ & 61.53$_{\pm1.35}$\\
    \midrule
         & \checkmark &  &  &  & 83.56$_{\pm7.92}$ & {80.53$_{\pm9.57}$} & \textbf{82.25$_{\pm14.96}$} & \underline{82.97$_{\pm15.21}$} & 69.71$_{\pm0.43}$ & 69.68$_{\pm0.54}$\\
         & \checkmark  &  & \checkmark & \checkmark & 84.15$_{\pm3.50}$ & 81.00$_{\pm5.59}$ & 73.98$_{\pm14.69}$ & 78.74$_{\pm12.19}$ & 70.29$_{\pm3.26}$ & \underline{71.22$_{\pm2.85}$} \\
         & \checkmark & \checkmark &  & \checkmark & 84.30$_{\pm7.40}$ & 82.52$_{\pm8.67}$ & 79.00$_{\pm14.58}$ & 82.68$_{\pm11.30}$ & 69.51$_{\pm3.19}$ & 70.48$_{\pm2.03}$ \\
         & \checkmark & \checkmark & \checkmark &  & 84.89$_{\pm5.53}$ & \underline{83.48$_{\pm7.03}$} & 74.63$_{\pm16.48}$ & 77.17$_{\pm14.66}$ & \underline{70.49$_{\pm1.76}$} & 70.82$_{\pm1.90}$ \\
        LCM & \checkmark & \checkmark & \checkmark & \checkmark & \textbf{86.30$_{\pm5.80}$} & \textbf{85.33$_{\pm7.35}$} & {81.30$_{\pm14.55}$} & \textbf{84.18$_{\pm11.63}$} & \textbf{71.46$_{\pm2.62}$} & \textbf{72.50$_{\pm1.91}$}\\
    \bottomrule
    \end{tabular}
    \label{tab:finetune}
    \vspace{-1em}
\end{table*}

\begin{table*}[t]
    \centering
    \setlength{\tabcolsep}{2pt}
    \small
    \caption{Performance on sex prediction across 7 datasets, where \textbf{Bold} indicates the first ranking place, and \underline{underline} indicates the second.}
    \vspace{-1em}
    \begin{tabular}{llllllll}
    \toprule
    Sex$\uparrow$           & HCPA       & HCPYA      & ADNI        & ABIDE       & PPMI        & Taowu       & Neurocon          \\
    \midrule
    BrainLM       & 40.68$_{\pm4.03}$ & 38.97$_{\pm5.40}$ & 41.72$_{\pm7.19}$  & \underline{78.59$_{\pm4.98}$}  & 42.03$_{\pm9.21}$  & 62.00$_{\pm23.53}$ & 49.33$_{\pm28.08}$ \\
    {BrainMass-SVM} & 69.44$_{\pm1.69}$ & 68.12$_{\pm3.56}$ & 45.40$_{\pm12.88}$ & 73.84$_{\pm3.49}$  & 48.83$_{\pm6.27}$  & 46.24$_{\pm23.96}$ & 33.24$_{\pm15.34}$ \\
    {BrainMass-MLP} & \underline{69.93$_{\pm1.64}$} & \underline{68.54$_{\pm3.36}$} & \underline{65.74$_{\pm7.66}$}  & 74.73$_{\pm4.20}$  & \underline{67.73$_{\pm12.68}$} & 69.29$_{\pm17.74}$ & 61.43$_{\pm27.90}$ \\
    {Brain-JEPA}     & 43.53$_{\pm0.65}$ & 43.63$_{\pm3.27}$ & 62.34$_{\pm6.51}$  & 78.11$_{\pm6.41}$  & 48.46$_{\pm7.09}$  & \textbf{92.48$_{\pm11.16}$} & \underline{83.33$_{\pm12.43}$} \\
    LCM    & \textbf{73.94$_{\pm2.45}$} & \textbf{72.23$_{\pm1.92}$} & \textbf{71.98$_{\pm6.87}$}  & \textbf{87.34$_{\pm4.48}$}  & \textbf{77.97$_{\pm3.76}$}  & \underline{90.67$_{\pm11.43}$} & \textbf{100.00$_{\pm0.00}$} \\
    \bottomrule
    \end{tabular}
    \label{tab:sexage}
\end{table*}

\begin{table*}[t]
    \centering
    \small
    \setlength{\tabcolsep}{2pt}
    \caption{Model scalability demonstration via performance on phenotypic prediction. \textbf{Bold} indicates the first ranking place, and \underline{underline} indicates the second.}
    \label{exp:tab-f1}
    \vspace{-1em}
  
\begin{tabular}{llllllll}
\toprule
 &  HCPA & HCPYA & ADNI & PPMI & ABIDE & Taowu & Neurocon \\
 & 3-task & 7-task & {Alzheimer's} & {Parkinson's} & {Autism} & {Parkinson's} & {Parkinson's} \\
\midrule
MLP-Small & 93.99$_{\pm0.35}$ & 88.67$_{\pm2.07}$ & 76.39$_{\pm8.90}$ & \underline{61.45$_{\pm11.05}$} & 68.81$_{\pm3.04}$ & 61.00$_{\pm28.29}$ & 65.87$_{\pm31.40}$\\
MLP-Mid & 91.87$_{\pm1.09}$ & 80.04$_{\pm3.40}$ & 75.85$_{\pm8.07}$ & 58.32$_{\pm15.46}$ & 66.14$_{\pm6.56}$ & 50.67$_{\pm29.09}$ & 62.48$_{\pm29.39}$\\
MLP-Big & 87.37$_{\pm8.25}$ & 76.43$_{\pm1.86}$ & 76.17$_{\pm7.36}$ & 41.74$_{\pm12.82}$ & 49.21$_{\pm14.90}$ & 43.00$_{\pm31.84}$ & 59.54$_{\pm30.60}$\\
\midrule
LCM-Small & 97.04$_{\pm0.40}$ & 94.63$_{\pm0.77}$ & 76.99$_{\pm9.91}$ & 57.29$_{\pm14.52}$ & 68.34$_{\pm3.54}$ & 74.57$_{\pm27.90}$ & 59.95$_{\pm27.60}$\\
LCM-Mid & \underline{97.15$_{\pm0.25}$} & \underline{94.84$_{\pm0.61}$} & \underline{77.96$_{\pm13.41}$} & 58.19$_{\pm13.35}$ & \underline{69.39$_{\pm5.02}$} & \underline{78.33$_{\pm20.39}$} & \underline{69.21$_{\pm28.19}$}\\
LCM-Big & \textbf{97.18$_{\pm0.54}$} & \textbf{95.02$_{\pm1.00}$} & \textbf{79.58$_{\pm13.77}$} & \textbf{61.61$_{\pm14.69}$} & \textbf{69.58$_{\pm5.23}$} & \textbf{88.33$_{\pm22.52}$} & \textbf{71.87$_{\pm27.42}$} \\
\bottomrule
\end{tabular}
\end{table*}

\subsection{Datasets}

We partition brain regions using the AAL atlas \cite{tzourio2002automated} through all experiments. The data preprocessing details can refer to the Supplementary and a benchmark paper \cite{xu2023data}.

\textbf{The Lifespan Human Connectome Project Aging (HCPA)} dataset \cite{bookheimer2019lifespan} is instrumental in task recognition research, offering a comprehensive view of the aging process. It includes data from 717 subjects, encompassing fMRI records ($n$=4,863) with human behaviors associated with memory, sensory-motor and the resting state. In our experiments, these tasks are treated as a four-class classification. 

\textbf{The Human Connectome Project Young Adult (HCPYA)} dataset \cite{van2013wu} has tackled key aspects of the neural pathways that underlie brain function and behavior via high-quality neuroimaging data in over 1100 healthy young adults. It includes data from seven human behaviors associated with various cognitive tasks, e.g., language and working memory. In our experiments, these tasks are treated as a seven-class classification. 

\textbf{Alzheimer’s Disease Neuroimaging Initiative (ADNI)} dataset \cite{weiner2015impact} serves as an invaluable resource, featuring a collection of pre-processed fMRI ($n$=138) and including clinical diagnostic labels. It encompasses a spectrum of cognitive states: Cognitive Normal (CN), Subjective Memory Complaints (SMC), Early-Stage Mild Cognitive Impairment (EMCI), Late-Stage Mild Cognitive Impairment (LMCI), and Alzheimer's Disease (AD). Considering the class unbalance issue, we simplified these categories into two broad groups based on disease severity: we combined CN, SMC, and EMCI into `CN' group, while LMCI and AD were grouped as the `AD' group.

\textbf{Parkinson’s Progression Markers Initiative (PPMI)} dataset \cite{xu2023data} presents a substantial collection of data from 209 subjects. It encompasses states of mental health: normal control, scans without evidence of dopaminergic deficit (SWEDD), prodromal, and Parkinson’s disease (PD). In our experiments, the dataset is treated as a four-class classification. 

\textbf{Autism Brain Imaging Data Exchange (ABIDE)} dataset presents data from 1025 young adults. The initiative aggregated fMRI data collected from laboratories around the world to support the research on Autism Spectrum Disorder (ASD). Subjects are classified into typical controls and those suffering from ASD. The binary classification is set for this dataset in our experiments. 

\textbf{Taowu} and \textbf{Neurocon} \cite{xu2023data} are two of the earliest image datasets released for Parkinson’s and contain 81 subjects. 
The binary classification is set for these datasets in our experiments.  

\textbf{Schizophrenia (SZ)} contains 189 subjects. There are 30 diseased and 159 healthy. The binary classification is set for the dataset.


\subsection{Implementation Details}

Following previous works, our experiments are done with subject-level cross-validation (CV). The average score and the standard deviation are both listed. To make our results comparable with previous papers, HCPA, HCPYA, and ADNI use a 5-fold CV as same as \cite{dan2023re,wei2024neuropath}, while others use 10-fold as same as \cite{xu2023data}. Since LCM is a foundation model, training data for pretraining and finetuning is always from the corresponding CV fold's training set to prevent data leakage. Hyperparameters, e.g., learning rate and hidden channels, can be found in the Supplementary.
SOTA brain-dedicated models, BrainGNN \cite{li2021braingnn}, BNT \cite{kan2022brain}, BolT \cite{bedel2023bolt}, and NeuroPath \cite{wei2024neuropath} are implemented as their original codes with default hyperparameters, where the structural connectome utilized by NeuroPath is replaced by FC in our work. Additionally, SOTA graph Transformers, Graphormer \cite{ying2021transformers}, and NAGphormer \cite{chen2022nagphormer}, are also compared. {The released BrainLM \cite{ortega2023brainlm}, the one we trained from scratch, along with the original BrainMass-SVM \cite{yang2024brainmass}, the modified BrainMass-MLP, and Brain-JEPA \cite{dong2024brain}, are both retrained and compared.} Codes and model weights can be found in the Supplementary.

\subsection{Main Results: Finetuning Performance}

\begin{table}
    \centering
    \small
    \setlength{\tabcolsep}{2pt}
    \caption{Fewshot finetuning (FT) performance comparison on a held-out disease.}
    \vspace{-1em}
    \label{exp:fewshot}
    \begin{tabular}{llll}
    \toprule
        Schizophrenia (SZ) & 10\% FT & 50\% FT & 100\% FT \\ \midrule
        BolT & 79.70$_{\pm9.76}$ & 80.16$_{\pm12.68}$ & 81.14$_{\pm9.75}$ \\ 
        NeuroPath & 78.60$_{\pm9.77}$ & 80.16$_{\pm12.68}$ & 81.74$_{\pm8.90}$ \\ 
        BrainMass-SVM & 78.60$_{\pm8.74}$ & 78.60$_{\pm8.74}$ & 78.60$_{\pm8.74}$ \\ 
        BrainMass-MLP & \underline{78.89$_{\pm10.14}$} & \underline{80.95$_{\pm11.00}$} & \underline{82.55$_{\pm10.95}$} \\ 
        LCM & \textbf{81.63$_{\pm8.58}$} & \textbf{81.73$_{\pm11.48}$} & \textbf{83.61$_{\pm6.01}$} \\ 
        $p$-value & 0.0267 & 0.0124 & 0.0014 \\ \bottomrule
    \end{tabular}
\end{table}

\begin{figure*}[t]
    \centering
    \includegraphics[width=\linewidth]{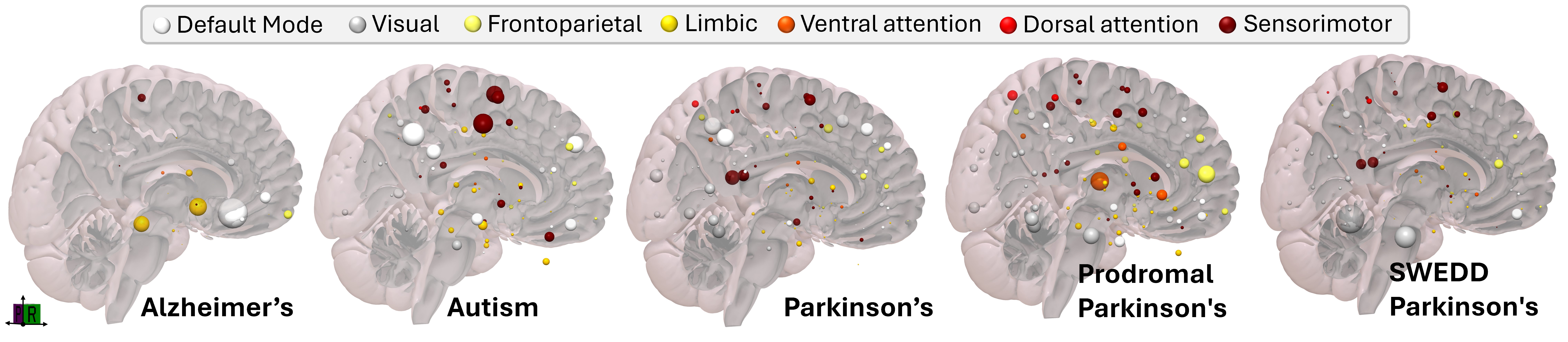}
    \vspace{-2em}
    \caption{The average cross-attention map of all test data at the readout layer of LCM on disease-related datasets. The node size indicates the relative attention weight.}
    \vspace{-.5em}
    \label{fig:attn-map}
\end{figure*}

\begin{figure}[t]
    \centering
    \includegraphics[width=\linewidth]{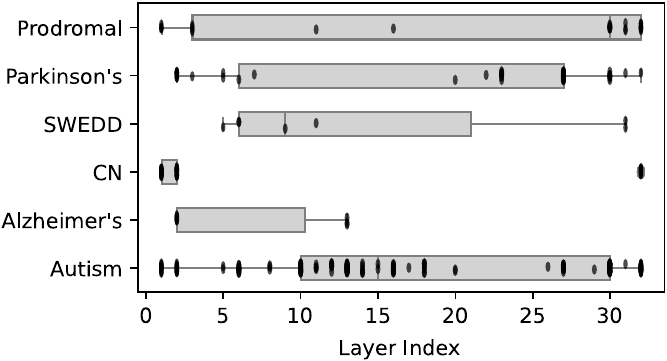}
    \vspace{-1.5em}
    \caption{The distribution of the index of the best-matched readout layer of LCM, where a dot in the box plot represents one sample point.}
    \label{fig:best_layer}
\end{figure}

\textbf{Disease diagnosis}
As listed in Table \ref{tab:finetune}, there are four additional versions of pretraining LCM, (i): without any diseases, (ii): no Alzheimer's, (iii): no Parkinson's, and (iv) no Autism, to compare with LCM pretraining with all data in the last row. SOTA models are also listed in the upper part of the Table for comparison. Note that, due to class unbalance, accuracy can have different ranks as F1 score, and F1 is the metric to conclude.

Firstly, pretraining with more data leads to better performance, and hence, the complete version of LCM has the best accuracy and F1 score against others. LCM pretrained with one or more disease-related datasets can outperform the one pretrained with only HCPs (\circled{1}) except for Parkinson's, which is the only four-class classification task and has higher difficulty.

Additionally, pretraining enhances the LCM performance even if the downstream tasks are unseen. After holding out all disease-related datasets (pretrained on \circled{1}), the F1 scores on all diseases are all better than the train-from-scratch LCM$\dagger$. Performance improvement is impressive in Parkinson's, given a 16.87\% increase in F1, while before pretraining LCM$\dagger$ is about 11\% lower than Brain-JEPA in F1. Compared to LCM$\dagger$, LCM is always better in F1. We can find this observation for cases of held-out Alzheimer's (pretrained on \circled{1} \circled{3} \circled{4}), Parkinson's (pretrained on \circled{1} \circled{2} \circled{4}), or Autism (pretrained on \circled{1} \circled{2} \circled{3}).

Last but not least, finetuning the LCM on seen datasets is more stable than on unseen datasets. Take Alzheimer's as an example, pretrained on \circled{1} \circled{2} \circled{3} and \circled{1} \circled{2} \circled{4} have 82.52\% and 83.48\% F1, respectively, while the best of unseen is 81\%. Autism also shows better when pretrained on \circled{1} \circled{3} \circled{4} than on \circled{1} \circled{2} \circled{3} in F1 scores. Whilst Parkinson's shows the opposite due to the difficult four-class classification.

\textbf{Demographics}
We test models to predict the sex on our seven datasets, respectively, as listed in Table \ref{tab:sexage}, where sex is measured by F1 score. 
Clearly, LCM outperforms others on all datasets except for Taowu ($n=40$), while other models are unable to hold the first/second place across all datasets. 

\subsection{Main Results: Fewshot Finetuning}

To further demonstrate the generalizability, we finetune and evaluate LCM on a new unseen dataset, SZ ($n=189$) by F1 score as listed in Table \ref{exp:fewshot},  $p$-value is from a paired t-test between scores of LCM and other SOTA models in the 5-fold cross-validation. As the experiments in BrainMass, different ratios of finetuning data, 10\%, 50\%, and 100\%, were used in the evaluation. We can see the generalizability of LCM is significantly better ($p<$0.05) than BrainMass.

\subsection{Main Results: Model Scalability}

The comparison of classification performance measured by F1 score is listed in Table \ref{exp:tab-f1}, where Small refers to 8-layer (100M-level), Mid is 20-layer (700M-level), and Big is 32-layer (1.2B-level). In contrast, SOTA models in the middle rows have only 4M-level parameters. Note that although either width or depth can increase the model scale of MLP, we found that width is not as valuable as the depth of MLP, referring to the similar accuracy with different widths as shown in the Supplementary. Except for BrainLM and LCM-Big which are single model pretrained with additional data and tested on each dataset, all models listed in Table \ref{exp:tab-f1} are trained from scratch for each dataset. 

Clearly, LCM holds the best/second performance given the best scalability, where the F1 score is always increased when the model size is enlarged, and it is finally significantly boosted by the pretraining. MLP is not scalable, agreeing with the exploded training loss as shown in Fig. \ref{fig:loss-nlayer}. The best performance by MLP is consistently demonstrated by the small version, while larger MLPs have lower F1 scores. For example, MLP-Small ranks in the top 5 for PPMI and ABIDE, but MLP-Big drops to last place. More analysis of MLP scalability can be found in Supplementary.

It is worth noting that LCM-Big got enhanced to the best performance with over 90\% F1 on Taowu and Neurocon datasets by pretraining (see Table \ref{tab:finetune}), while before that, LCM-Big was out of the top three. LCM drops on HCPA and HCPYA after pretraining because their sample size is 10 times the rest of the datasets. However, the disease-related datasets that have small sample sizes consistently benefited from big data.

\subsection{Interpretation}

{\textbf{Adaptive training}  We demonstrate the frequency of the best-matched readout layer index across batches by box plot in Fig \ref{fig:best_layer}. We can see it has a diverse distribution for various disease diagnoses. Alzheimer's, Control Normal (CN), and SWEDD tend to use shallow features at layers between 1 to 15 across batches since it is relatively easier to separate Alzheimer's (dementia stage) and CN/SWEDD (with non-evident symptoms). In contrast, multi-level feature representations are required from various layers to effectively differentiate Autism, Parkinson's, and Prodromal stages because of subtle variations in brain function. This distribution is collected from the last epoch of finetuning, indicating semi-supervised LCM did not converged to a fixed layer for different phenotypes.}

\textbf{Visualizations}  The average cross-attention map of test data with the same label in ADNI, ABIDE, and PPMI datasets is shown in Fig. \ref{fig:attn-map}. The attention weights are extracted at the readout layer by the proposed adaptive training. We can observe that LCM is more attentive to the default mode network for Alzheimer's and Autism than Parkinson's, which aligns with current neuroscience knowledge \cite{padmanabhan2017default,zhang2023dissociable}. It is clear that LCM is also attentive to the limbic network for ADNI and ABIDE, which is often damaged in Alzheimer's disease \cite{hopper1976limbic} and Autism \cite{wong2020serotonin}. For three classes of Parkinson's, LCM is attentive to sensorimotor, visual, and frontoparietal networks that are involved in Parkinson's disease by agreeing with \cite{schneider1987parkinson,cascone2021frontoparietal,gottlich2013altered}. These visualizations can interpret the promising performance of LCM.

\begin{table*}[h]
    \centering
    \small
    \setlength{\tabcolsep}{2pt}
    \caption{Ablation studies of LCM learning strategy, where average scores are listed.}
    \vspace{-1em}
    \begin{tabular}{llll|llllll}
    \toprule
         & \multicolumn{3}{c}{Supervised at} & \multicolumn{2}{c}{Alzheimer’s} & \multicolumn{2}{c}{Parkinson’s} & \multicolumn{2}{c}{Autism}\\\cmidrule(lr){2-4}\cmidrule(lr){5-6}\cmidrule(lr){7-8}\cmidrule(lr){9-10}
         & Last & Best & All & Accuracy & F1 score & Accuracy & F1 score & Accuracy & F1 score \\
    \midrule
        Baseline & \checkmark & &  & 83.22$_{\pm5.72}$ & 80.49$_{\pm6.69}$ & 76.06$_{\pm16.02}$ & 78.45$_{\pm14.31}$ & \underline{69.35$_{\pm3.22}$} & \underline{69.90$_{\pm3.34}$} \\
        Only stage 1 &  &  & \checkmark  & 79.19$_{\pm6.33}$ & 80.49$_{\pm6.63}$ & \underline{80.40$_{\pm11.13}$} & \textbf{84.74$_{\pm8.35}$} & 65.24$_{\pm2.86}$ & 66.41$_{\pm2.25}$\\
        Only stage 2 &  & \checkmark &  & \underline{84.15$_{\pm3.50}$} & \underline{83.78$_{\pm6.69}$} & 75.35$_{\pm13.80}$ & 80.95$_{\pm7.68}$ & 66.80$_{\pm2.10}$ & 67.32$_{\pm1.76}$ \\
        Stage 1 \& 2 &  & \checkmark & \checkmark & \textbf{86.30$_{\pm5.80}$} & \textbf{85.33$_{\pm7.35}$} & \textbf{81.30$_{\pm14.55}$} & \underline{84.18$_{\pm11.63}$} & \textbf{71.46$_{\pm2.62}$} & \textbf{72.50$_{\pm1.91}$}\\
    \bottomrule
    \end{tabular}
    \label{tab:ablation}
\end{table*}

\subsection{Ablation studies}

As introduced in Section \ref{sec:method}, the learning strategy of LCM is designed with a momentum of full supervision (stage 1) on all decoder layers in a few beginning epochs, and then followed by an adaptive training (stage 2) on the best-matched layer. To show the effectiveness of the learning strategy design, we run finetuning experiments for the pretrained LCM with all three ablate version as listed in Table \ref{tab:ablation}, where the model supervised at the last layer as the baseline represents the training of a generic Transformer decoder. Only stage 2 across the entire finetuning shows the necessity of the momentum of full supervision at stage 1, and only stage 1 shows the necessity of stage 2.

Clearly, as listed in Table \ref{tab:ablation}, our design of training LCM shows the best performance on disease-related datasets. Other versions of LCM finetuning are not as stable as the performance gained by our design. Although the only stage 2 version shows the second-best performance on Alzheimer's, it has worse scores than the train-from-scratch LCM on Autism (see Table \ref{tab:finetune}). As well as the only stage 1 version acts well on Parkinson's but fails on Autism. This observation supports that different phenotypes perform great at different layers of LCM, and momentum by full supervision is required for training LCM on the correct direction.

\section{Conclusion}
\label{sec:conc}

In conclusion, we proposed a large connectome model (LCM), which is the {largest} brain foundation model (1.2B) for clinical applications. Tremendous efforts have been made to pretrain large scalable models on extensive fMRI data using self-regressive learning. Even though every fMRI data has phenotypes or demographics recorded, none has explored a multitask methodology for the large brain foundation model. By capitalizing on rich environmental variables and demographic data along with an unprecedented amount of fMRI data, we present a scalable model architecture for more effective pretraining by multitask learning for brain-environment interactions (BEI). We design the LCM for scalability and the robustness of finetuning with the consideration of the diverse complexity of feature representation for different BEIs. We have evaluated our foundation model on a variety of applications, including sex prediction, human behavior recognition, and disease early diagnosis of Autism, Parkinson’s disease, and Alzheimer’s disease, where promising results shown by the pretrained LCM indicate the great potential to facilitate brain connectome in clinical routines. The LCM finetuning on unseen datasets is also promoted by the pretraining with significant performance enhancement compared to train-from-scratch LCM. Given the impressive performance of our methods on 8 datasets, the decoder-only architecture learning from the multitask learning provides a new routine for training a brain foundation model.

\section*{Acknowledgement}

This work was supported by the National Institutes of Health (AG091653, AG068399, AG084375) and the Foundation of Hope.

\bibliography{aaai2026}

\newpage
\appendix

\onecolumn
\section{Accessibility}

Public data is accessible via internet (HCPA\footnote{\url{https://www.humanconnectome.org/}}, HCPYA\footnote{\url{https://www.humanconnectome.org/study/hcp-young-adult/overview}}, ADNI\footnote{\url{https://adni.loni.usc.edu/}}. PPMI, ABIDE, Taowu, and Neurocon can be found here\footnote{\url{https://auckland.figshare.com/articles/dataset/NeurIPS_2022_Datasets/21397377}}). The licenses to obtain those data can also be accessed on the websites.
The codes and data split settings can be acquired via this code repository\footnote{\url{https://github.com/Chrisa142857/brain_network_decoder}}. The weights of LCM-Big can be found here\footnote{\url{https://shorturl.at/3rYjK}}. Note that separate model weights were picked according to the best validation score on eight datasets, and the model weights used in the main text were picked by the HCPA score.

\section{Data profiles}

The data profiles of our seven datasets are listed in Table \ref{tab:data-profile}.

\begin{table*}[h]
    \centering
    \caption{The profiles of datasets used in this work.}
    \begin{tabular}{llllllll}
    \toprule
         & HCPA & HCPYA & ADNI & PPMI & ABIDE & Taowu & Neurocon \\
    \midrule
        Subject \# & 713 & 248 & 138 & 209 & 1,025 & 40 & 41\\
        FC \# & 4,863 & 3,293 & 138 & 209 & 1,025 & 40 & 41\\
        Class \# & 4 & 7 & 2 & 4 & 2 & 2 & 2 \\
        Sex ratio (M:F) & 0.79 & 0.84 & 0.89 & 1.55 & 5.68 & 1.35 & 0.86\\
        Age range & [36, 100] & [22, 36] & [55, 89.6] & [38, 84] & [6, 58] & [57, 75] & [45, 86]\\
    \bottomrule
    \end{tabular}
    \label{tab:data-profile}
\end{table*}


\section{Data preprocessing}

\begin{figure*}[b]
    \centering
    \includegraphics[width=\textwidth]{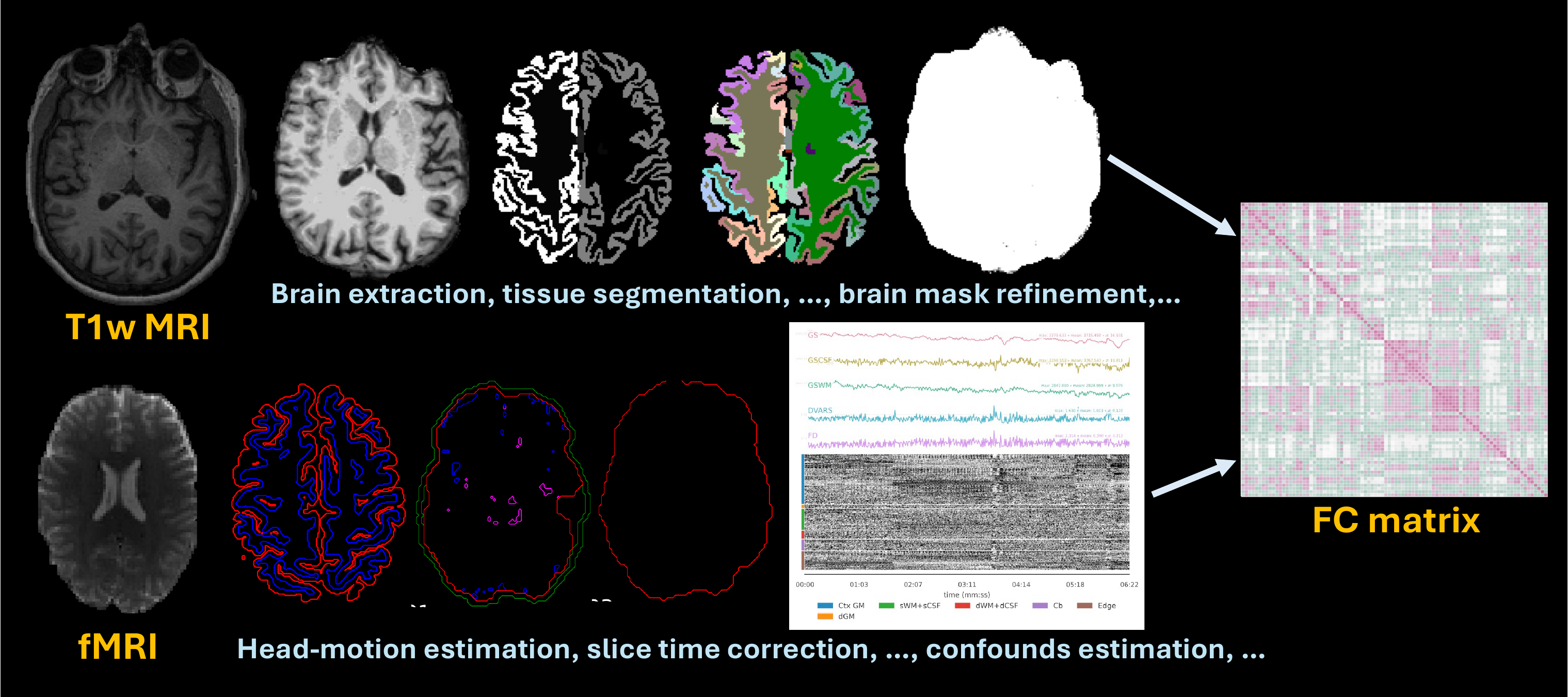}
    \caption{General workflows for processing T1-weighted image (T1w MRI) and functional MRI (fMRI). The output is shown at the right, including the brain network of FC.}
    \label{fig:data_prep}
\end{figure*}

The neuroimage processing consists of the following major steps:  
(1) We segment the T1-weighted image into white matter, gray matter, and cerebral spinal fluid using FSL software \cite{FSL}. (2) On top of the tissue segmentation in Fig. \ref{fig:data_prep}, we parcellate the cortical surface of fMRI into cortical regions according to the atlas as a regional signal of time-series in Fig. \ref{fig:data_prep}, where FC, in the end, is the Pearson correlation coefficient between regional time-series. 

\section{{Differences with other brain foundation models}}
We propose to develop a foundation model of fMRI using scalable logistic regression. The promising performance by our LCM proves the effectiveness of this novel idea. In addition, our LCM is highly different from others as listed in Table \ref{tab:difference}.

\begin{table*}[!ht]
    \centering
    \caption{Differences with other brain foundation models}
    \label{tab:difference}
    \begin{tabular}{ccccc}
    \toprule
        ~  & BrainLM & BrainMass & BrainJEPA & LCM \\ \midrule
        Input   & BOLD & FC & BOLD & FC  \\ 
        Model scale  & 650M & 67M & 307M & 1.2B  \\ 
        Model architecture  & ViT & Encoder-only & ViT & Decoder-only  \\ 
        Pretrain   & BOLD recon & FC recon & Latent feature recon & Brain-env interaction  \\ 
        \thead{Downstream \\Phenotypes}  & \thead{Age, PTSD,\\ Anxiety, \\Neuroticism} & - & \thead{Age, Sex, \\Neuroticism, \\Flanker, Amyloid} & Sex, Tasking \\ 
         Diseases  & -  & \thead{ASD, ADHD, \\AD, PD, MDD}  & AD  & ASD, AD, PD, SZ   \\ \bottomrule
    \end{tabular}
\end{table*}

\section{{Computational cost}}
Our two-stage method makes the internal evaluation a bit more complex than self-supervised methods. The reason is that we need to exclude the evaluation dataset from LCM pre-training to prevent data leakage. In contrast, self-supervised methods pre-train with all data, including the evaluation dataset. However, for evaluating external diseases, such as Schizophrenia (SZ) in the main text, the evaluation procedure of LCM is the same as self-supervised methods because they are absolutely unseen for either self- or semi-supervised methods. 

In terms of actual training time used for pre-training and fine-tuning, we list the time cost per sample in Table \ref{tab:time_cost} along with parameter numbers. Self-supervised methods, BrainJEPA and BrainMass, build additional neural networks at fine-tuning leading to more time cost. In contrast, our LCM only fine-tunes additional tokens leading to less computational cost as pre-training. In total, our LCM requires the least computational cost with the most learnable parameters.

\begin{table*}[!ht]
    \centering
    \caption{Time cost of brain foundation models}
    \label{tab:time_cost}
    \begin{tabular}{lllll}
    \toprule
        ~ & BrainLM & BrainJEPA & BrainMass & LCM  \\ \midrule
        Pretrain time (ms) & 22.32 & 4.82 & 3.75 & 13.09  \\ 
        Finetune time (ms) & 6.68 & 28.13 & 37.08 & 12.04  \\ 
        Param. \# & 650M & 307M & 67M & 1.2B  \\ \bottomrule
    \end{tabular}
\end{table*}

\section{Computing environments and hyperparameters}

The experiments are done on a Linux system with one NVIDIA RTX 6000 Ada. Batch size and learning rate are set as 128 and 1e-4, respectively. The maximum epoch is set as 200 and $m=5$. $D=E=2048$. Training will be early stopped if accuracy keeps dropping in 50 epochs. SOTA models including BrainGNN, BolT, BNT, Graphormer, NAGphormer, NeuroPath, and the finetuning head of BrainLM are all using the same number of hidden channels, $D=2048$, to meet the same level of feature representation. BrainLM is pre-trained on our datasets with the default hyperparameters, where the time-series is cropped into 30 time points during the pre-training.

\section{MLP scalability analysis}

\begin{figure*}[h]
    \centering
    \includegraphics[width=\linewidth]{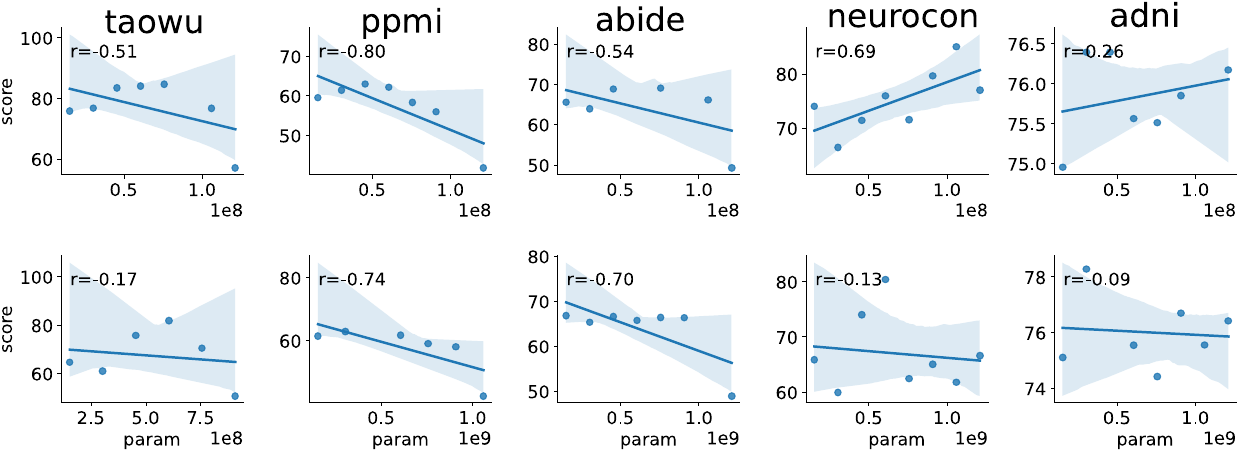}
    \vspace{-2em}
    \caption{The scalability of MLP on disease prediction using 2048 hidden channels (first row) and 6177 hidden channels (second row), where the y-axis is the F1 score.}
    \label{fig:mlp-scale}
\end{figure*}

MLP as a universal predictive head is used in most related works for downstream applications. In this work, we test the scalability of MLP on the same experiments as we run for the LCM. Fig. \ref{fig:mlp-scale} is the comparison between MLPs using 2048 hidden channels and 6177 hidden channels with different layer numbers.

\end{document}